%% file: neurips_2025.tex
\documentclass{article}

     \usepackage[preprint]{neurips_2025}

\usepackage{array}       

\usepackage{amsmath}     
\usepackage{amssymb}     
\usepackage{multirow}    

\usepackage[dvipsnames]{xcolor}

\usepackage{makecell}

\usepackage{amsmath}%

\usepackage{MnSymbol}%
\usepackage{wasysym}%

\usepackage[utf8]{inputenc} 
\usepackage[T1]{fontenc}    
\usepackage{hyperref}       
\usepackage{url}            
\usepackage{booktabs}       
\usepackage{amsfonts}       
\usepackage{nicefrac}       
\usepackage{microtype}      
\usepackage{xcolor}         
\usepackage{colortbl}
\usepackage{graphicx}
\usepackage{cleveref}
\definecolor{LightGrey}{rgb}{.9,.9,.9}
\definecolor{White}{rgb}{1.,0.,1.}
\definecolor{first}{rgb}{.8,.0,.0}
\definecolor{second}{rgb}{.0,.6,.0}
\definecolor{third}{rgb}{.0,.0,.8}

\usepackage{pifont}
\usepackage{booktabs}
\usepackage{multirow}
\usepackage{amssymb} 
\usepackage{color} 
\usepackage{colortbl} 
\usepackage{subcaption}

\newcommand{\sysname}{SHTOcc }

\title{SHTOcc: Effective 3D Occupancy Prediction with Sparse Head and Tail Voxels}

\author{
    Qiucheng Yu \\ 
    City University of Hong Kong\\ 
  \texttt{qiuchenyu2-c@my.cityu.edu.hk} \\ 
  \And 
  Yuan Xie \\ 
  East China Normal University \\ 
  \texttt{yxie@cs.ecnu.edu.cn} \\ 
   \And 
  Xin Tan\thanks{Corresponding author} \\ 
  East China Normal University \\ 
  \texttt{xtan@cs.ecnu.edu.cn} \\ 
}

\begin{document}

\maketitle

\begin{abstract}

3D occupancy prediction has attracted much attention in the field of autonomous driving due to its powerful geometric perception and object recognition capabilities. However, existing methods have not explored the most essential distribution patterns of voxels, resulting in unsatisfactory results. This paper first explores the inter-class distribution and geometric distribution of voxels, thereby solving the long-tail problem caused by the inter-class distribution and the poor performance caused by the geometric distribution. Specifically, this paper proposes SHTOcc (\textbf{S}parse \textbf{H}ead-\textbf{T}ail Occupancy), which uses sparse head-tail voxel construction to accurately identify and balance key voxels in the head and tail classes, while using decoupled learning to reduce the model's bias towards the dominant (head) category and enhance the focus on the tail class. Experiments show that significant improvements have been made on multiple baselines: SHTOcc reduces GPU memory usage by 42.2\%, increases inference speed by 58.6\%, and improves accuracy by about 7\%, verifying its effectiveness and efficiency. The code is available at 
\href{https://github.com/ge95net/SHTOcc}{https://github.com/ge95net/SHTOcc}.

\end{abstract}

\input{chapter/introduction}
\input{chapter/related_work}
\input{chapter/methods}
\input{chapter/experiment}

\input{chapter/ablation_study}
\input{chapter/conclusion}

\bibliography{NeuIPS_sample}
\bibliographystyle{IEEEtran}

\appendix


\end{document}

%% file: chapter/introduction.tex
\section{Introduction}

Vision-based 3D occupancy prediction (3D Occ) constitutes a fundamental task in 3D scene understanding, requiring simultaneous estimation of occupancy states and semantic labels for each voxel within a volumetric space. This capability forms a critical component for robotic navigation and autonomous driving systems~\cite{behley2019semantickitti,wang2023openoccupancy,caesar2020nuscenes}, necessitating the development of computationally efficient feature processing methods to ensure reliable predictions. The inherent structural regularity of 3D voxel distributions provides valuable prior knowledge that remains underutilized in current approaches.

Contemporary methods fall into two primary categories: (1) \textit{Distribution-independent methods} approaches, including image-based implementations \cite{ming2024inversematrixvt3d,wei2023surroundocc,zhang2023occformer,hou2024fastocc,lu2024fast,tian2024mambaocc,yu2023flashocc} and video-based extensions \cite{ma2024cotr,chen2025rethinking,liao2025stcocc}, process complete 3D volumes through specialized decoders for per-voxel classification. (2) \textit{Partial Distribution-Aware Methods} techniques such as TPV \cite{huang2023tri,zhang2024lightweight,liang2024former}, 3D Gaussian Splatting (3DGS) \cite{huang2024gaussianformer,jiang2024gausstr,huang2024probabilistic}, and SparseOcc \cite{wang2024opus,liu2024fully,tang2024sparseocc,oh20253d} exploit voxel sparsity by focusing computations on occupied regions. While these methods demonstrate progressive improvements, they inadequately exploit two fundamental distribution patterns: (a) inter-class distribution and (b) geometric distribution. Our work systematically addresses both aspects through explicit distribution-aware modeling.

The inter-class distribution of voxels describes the varying frequency distribution across different object classes. We observe a pronounced long-tail distribution problem in 3D occupancy datasets, where head classes (e.g., roads and vegetation) may outnumber tail classes (e.g., bicycles and motorcycles) by several orders of magnitude. This severe class imbalance leads to biased predictions, as models tend to overfit to frequent classes while underperforming on rare ones. While the long-tail problem has been extensively studied in image classification~\cite{alshammari2022long,wang2020devil,kang2019decoupling}, it remains largely unaddressed in the context of 3D occupancy prediction. Furthermore, voxels exhibit inherent geometric distribution patterns. In 3D scenes, objects of the same type often occupy a large continuous space. For example, a road is usually a whole flat area. However, even when facing the same area, the existing method needs to check whether each voxel belongs to the same category. This repetitive work wastes time and computing power, and may even lead to wrong predictions.

\begin{figure}[t]
\centering
    \begin{subfigure}[b]{0.49\linewidth}
         \centering
         \includegraphics[width=\linewidth]{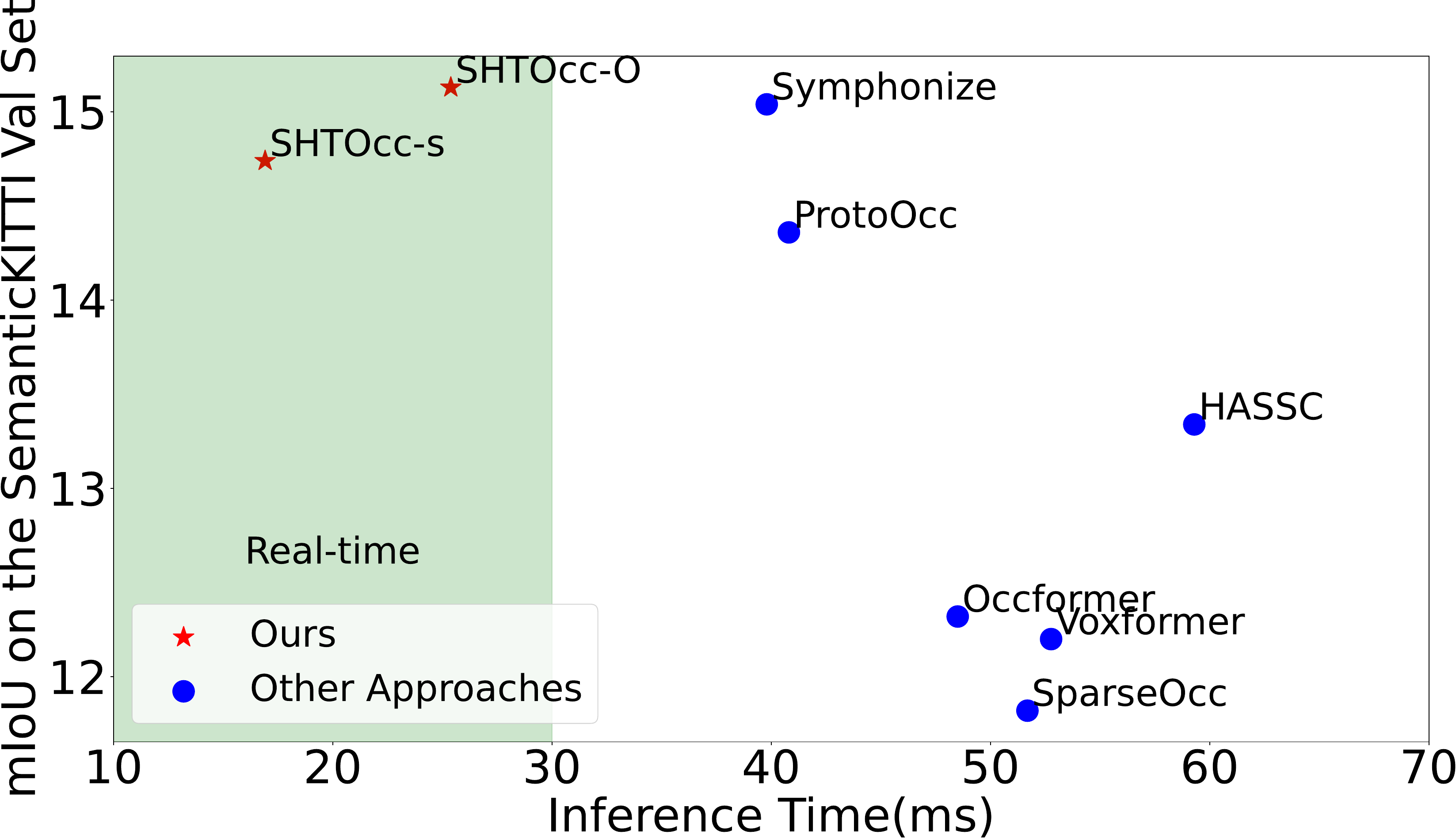}

         \caption{Comparisons of mIOU and inference time.}
         \label{time}
    \end{subfigure}
    \hfill
    \begin{subfigure}[b]{0.49\linewidth}
         \centering
         \includegraphics[width=\linewidth]{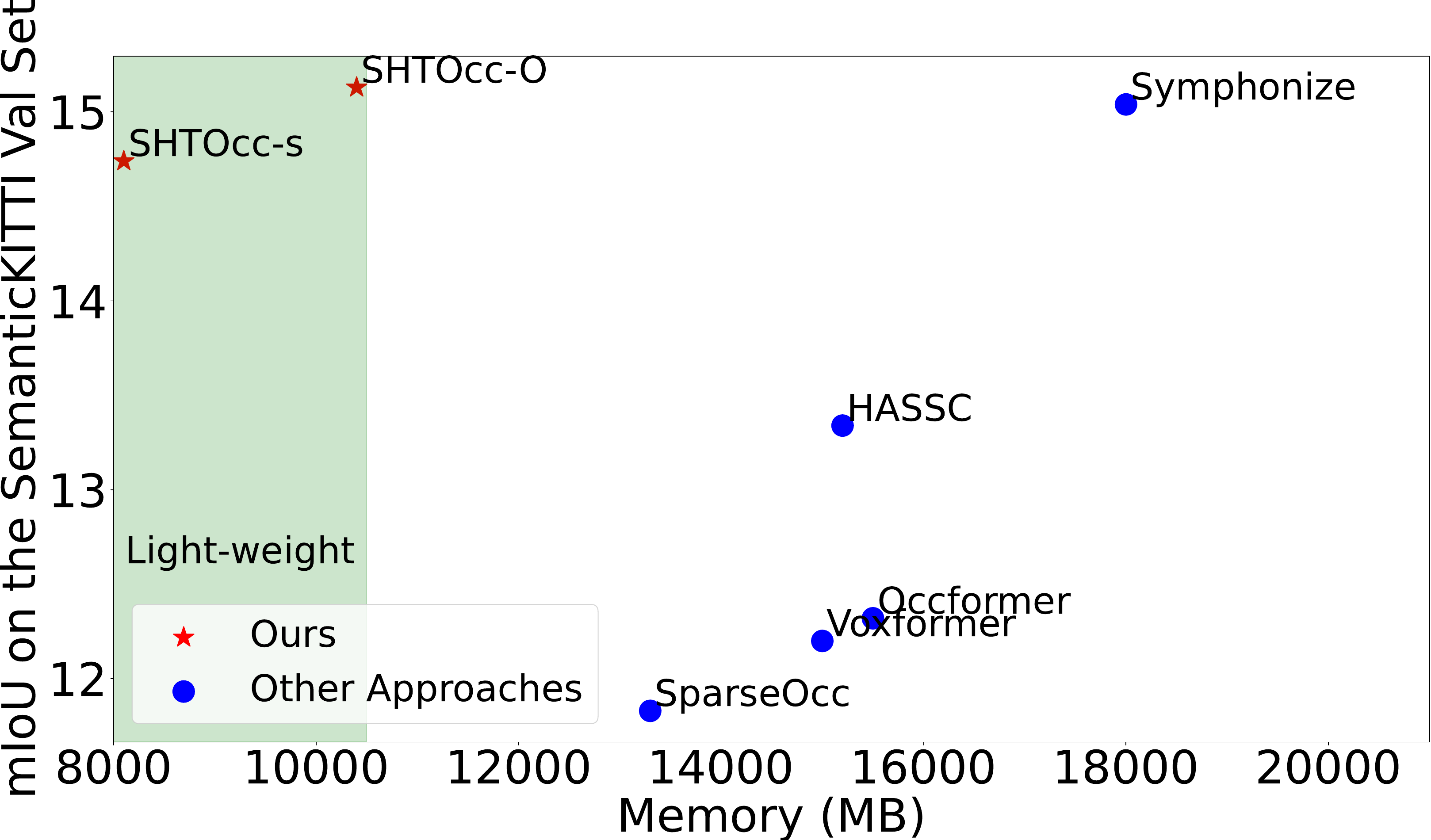}
  
         \caption{Comparisons of mIOU and training memory.}
         \label{memory}
    \end{subfigure}
    \begin{subfigure}[b]{\linewidth}
         \centering
         \includegraphics[width=1.\linewidth]{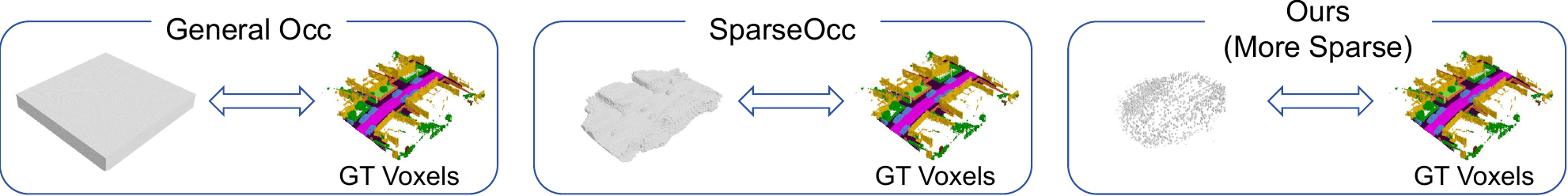} 
         \vspace{-0.2in}
         \caption{Key voxels selection. SpaseOcc selects all non-empty voxels, while \sysname lets the model select the most important voxels, which can achieve higher prediction accuracy with sparser voxels. } 
         \label{fig:key_voxels}
    \end{subfigure}

    \caption{Comparisons of the SHTOcc of various 3D semantic scene completion methods on the SemanticKITTI~\cite{behley2019semantickitti} dataset.}
    \label{fig:comparison}
\end{figure}

In this paper, we propose a new framework SHTOcc, which explores the inter-class distribution and geometric distribution of 3D voxels for the first time, and uses this as prior knowledge to guide the generation of voxels. 

\textbf{Sparse Head-Tail-Voxel Construction.} First, to avoid redundant recomputation of homogenized voxels, we propose a novel sparse sampling strategy to construct a very sparse voxel distribution that can correctly describe the geometric distribution of the spatial scene. Specifically, \sysname adopts a two-stage selection: (1) attention-guided top-k head voxel extraction to capture the key areas that the model itself pays most attention to; (2) balanced tail voxel sampling using class-specific confidence to extract tail class voxels ignored by the model. This dual mechanism explicitly retains the class information of the head and tail while maintaining computational efficiency.

\textbf{Decouple Decoder.} To address the long-tail problem caused by the imbalanced inter-class distribution, we introduced decoupled learning into 3D Occ named decouple decoder, which separates learning into two distinct phases: (1) Representation learning: the model is trained to learn a general feature representation without class distribution bias; (2) Class-balanced head refinement: Freeze the learned representations and exclusively retrain the segmentation head with label smoothing, strategically redistricting prediction confidence to prevent overfitting on head classes.


Through extensive experiments on the various task including Lidar Segmentation in nuScenes~\cite{caesar2020nuscenes}, 3D Occupancy prediction in nuScenes-Occupancy~\cite{wang2023openoccupancy} and Occ3D-nuScenes~\cite{tian2023occ3d}, Semantic Scene Completion SemanticKITTI~\cite{behley2019semantickitti} benchmarks, we verify that the proposed \sysname is an effective solution which can reduce memory overhead, improve inference speed, achieves the real time result and obtain accurate 3D Occ results. \Cref{time} and \Cref{memory} demonstrate that \sysname reduces inference time, saves computational memory, and improves prediction accuracy. \Cref{fig:key_voxels} demonstrates that the voxels of \sysname is much more sparse than SparseOcc ~\cite{tang2024sparseocc}.
The contributions are summarized as follows:

\begin{itemize}
    \item We propose a sparse head-tail voxel construction to carefully design sparse voxels, which can accurately extract key voxels while avoiding the model's excessive focus on head voxels and alleviating the long-tail effect.
    \item We introduce decoupled training with label smoothing into 3D Occ, which can further prevent the model from focusing too much on the head voxels and improve the prediction accuracy.
    \item Our method has been embedded in several popular backbone networks, and extensive experiments on various datasets show that our method significantly reduces the memory overhead while improving performance, achieving state-of-the-art performance.
\end{itemize}

    

%% file: chapter/related_work.tex
\section{Related Work}
\subsection{3D Occupancy Prediction}
\textbf{Distribution-independent methods.} The field of 3D semantic occupancy prediction has evolved significantly since its inception~\cite{mescheder2019occupancy,peng2020convolutional,song2017semantic}. Early approaches primarily utilized volumetric TSDF representations processed through 3D convolutions~\cite{li2019rgbd,zhang2019cascaded}, while subsequent hybrid methods~\cite{guo2018view,li2020anisotropic,liu2018see} focused on projecting 2D features into 3D space. Recent camera-based methods have introduced several innovations: MonoScene~\cite{cao2022monoscene} pioneered purely visual solutions with 3D UNets, TPVFormer~\cite{huang2023tri} developed tri-perspective view representations, and VoxFormer~\cite{li2023voxformer} introduced diffusion-based strategies. Other notable advances include mask-wise prediction in OccFormer~\cite{zhang2023occformer}, geometric enhancements in OccDepth~\cite{miao2023occdepth} and NDCScene~\cite{yao2023ndc}, and unified scene modeling in OccNet~\cite{tong2023scene}. While these methods demonstrate progressive improvements, they fundamentally treat all voxels equally, ignoring two critical distribution patterns: (1) the inherent sparsity of 3D scenes where most voxels are empty, and (2) the non-uniform semantic distribution across different object categories.

\textbf{Biased Distribution-Aware Methods.} Recent methods exploit voxel sparsity by focusing computation on occupied regions. (1) BEV-based methods ~\cite{ming2024inversematrixvt3d,hou2024fastocc,yu2023flashocc} compress dense voxels into BEV views, saving space but losing height information. (2) Sparse representation methods, such as TPV~\cite{huang2023tri,zhang2024lightweight} use three views to replace dense voxels, SparseOcc~\cite{tang2024sparseocc} prune empty voxels and only keep non-empty voxels, and ProtoOcc~\cite{oh20253d} use low-resolution queries to replace high-resolution queries. However, existing methods fail to fully utilize the inter-class semantic distribution and geometric regularity in the voxel space, resulting in poor performance. This paper explores and solves the related problems caused by the inter-class distribution and geometric distribution, further improving the prediction accuracy.


\subsection{Long Tail Distribution}
The long-tail recognition (LTR) problem has been extensively studied in various 2D computer vision tasks, including image classification~\cite{alshammari2022long}, object detection~\cite{wang2020devil}, and semantic segmentation. Popular solutions typically involve class re-weighting~\cite{alshammari2022long}, re-sampling strategies~\cite{wang2020devil}, or the decoupled training paradigm~\cite{kang2019decoupling}. Recent works like LTWB~\cite{alshammari2022long} and LOS~\cite{sunrethinking} have further advanced the field by analyzing the separate effects of representation learning and classifier retraining.

However, Traditional 2D task methods cannot be directly applied to 3D tasks, because 3D data has higher dimensional complexity (such as voxel sparsity and geometric structure sensitivity), simply resampling the tail class voxels will cause the geometric relationship of the 3D scene to be disordered. While long-tail problems have been addressed in many vision domains, no prior work has specifically targeted this issue in 3D occupancy prediction. This paper bridges this gap by adapting decoupled training and label smoothing techniques to 3D Occ, marking the first dedicated effort to solve the long-tail distribution problem in this domain.



%% file: chapter/methods.tex
\section{Methods}
\begin{figure*}[t]
    \centering
    
    \includegraphics[width=\linewidth]{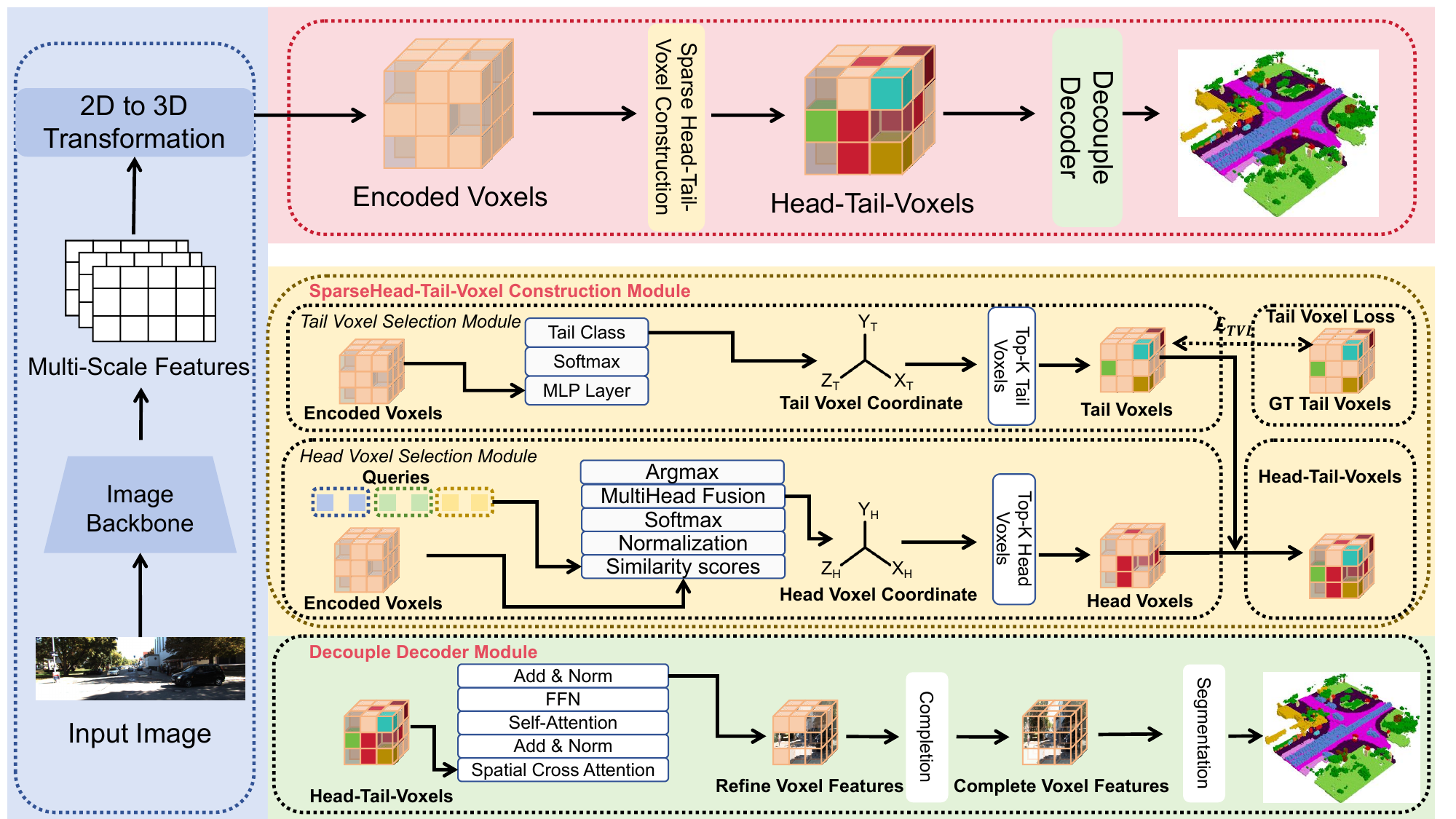}

    \caption{Overview of \sysname. The images featured are initially extracted by the image backbone and then convert to 3D encoded voxels through 2D to 3D transformation. The encoded voxel is extracted through dual path to obtain sparse Head and Tail voxels. The entire extraction process is a corase-to-fine process. With each additional layer, more voxels will be extracted accordingly. In the second learning phase of decouple training, only the parameters of segmentation head will be updated.}
    \label{fig:system_overview}

\end{figure*}
\subsection{Preliminary}

\textbf{3D Occupancy Prediction.} Given a monocular image or a set of images taken by surrounding cameras $I={I^{N}_{i}}$, the goal of vision-based 3D semantic occupancy prediction is to estimate the state of 3D voxels within the visual range and obtain the dense semantic voxel predictions $V\in \mathbb{R}^{C \times H\times W\times D}$, i.e., to predict the semantic label of each voxel in a predefined 3D voxels based on the known camera intrinsic parameters $K_{i}$ and extrinsic parameters ${[R_{i}|{t_{i}]}}$, where $N$ indicates the number of view cameras, $C$ is the number of classes and $H,W,D$ represents the voxel spatial resolutions.

\textbf{Sparse Representation.} According to the observation from SparseOcc~\cite{tang2024sparseocc,liu2024fully}, about $90\%$ of the voxels in $V$ are empty. Updating the entire voxel requires a lot of time and memory overhead. Therefore, sparse voxel representation is proposed to replace the dense voxel representation to avoid the waste of computing resources. Specifically, dense feature V are converted to a sparse representation by gathering non-empty voxels and the sparse tensor is stored in a commonly used coordinate (COO) format:

\begin{equation}
    \mathbb{V} = \{(\mathbf{p}_i = [x_i, y_i, z_i] \in \mathbb{R}^3, \mathbf{f}_i \in \mathbb{R}^C) | i = 1, 2, \ldots, N \}.
\end{equation}

In the above equation, $N$ represents the number of nonempty voxels, while $p_{i}$ and $f_{i}$ denote the coordinates and features of the $i-th$ voxel, respectively, all subsequent operations are performed on this sparse representation.

\subsection{Method Overview}
As shown in \Cref{fig:system_overview}, we process monocular or multi-view images $I={I^{N}_{i}}$ through three key stages: First, the multi-scale features  $F={F^{N}_{i}}$ are extracted by 2D image backbone. Then the encoded voxels are obtained through the 2D to 3D transformation module according to the multi-scale features and initial voxel queries. Second, The encoded voxels are passed through the Sparse Head-Tail-Voxel Construction module to leave the most important sparse head and tail voxels, and these head and tail voxels interact through the 3D backbone to get the preliminary prediction results. Finally, all parameters except the segmentation head are frozen, and decouple learning is performed to get the final output 3D semantic voxels.

\subsection{Sparse Head-Tail-Voxel Construction}
\label{Sparse Head-Tail-Voxel Construction.}



\textbf{Head Voxel Selection.} In the 3D voxel decoder, a key insight is that only a few important voxels actually need to be updated - but crucially, these important voxels should be determined by the model itself, not by manual observation. This motivates us to use the attention mechanism to automatically identify and prioritize the most task-relevant voxels. The attention mechanism dynamically computes the relevance score between the learned query ($Q$) and the voxel-based key ($K$), generating an attention matrix of shape ${L \times S}$ where each element represents the normalized importance weight ($0-1$) between the query and the voxel location. Through this process, the model is able to intrinsically learn which voxels are most important for occupancy prediction, as evidenced by their attention weights, eliminating the need to manually set importance criteria, such as non-empty voxels.

As shown in the yellow box in \Cref{fig:system_overview}, in 3D occupancy prediction tasks, voxel embeddings typically act as keys, while queries are strategically designed according to specific methods. We first measures query-key similarity through dot products $Q\cdot K^T$, producing raw attention scores that quantify spatial relationships across the 3D voxels. These scores then undergo normalization by scaling with $ \frac{1}{\sqrt{d_k}}$ to maintain stable gradients in the high-dimensional feature space, where $d_k$ is the feature dimension, followed by row-wise softmax to convert them into probabilistic attention weights that sum to 1 per query, ensuring focused attention on relevant regions. After that, the normalized attention from multiple parallel heads are aggregated through averaging to obtain the attention matrix $M_{atten} \in \mathbb{R}^{L\times S}$, where $L$ denotes the length of queries and $S$ is the length of keys. The attention matrix $M_{atten}$ quantifies the attention weights between all query-key pairs. By taking the argmax along the $0th$ dimension of $M_{atten}$, we identify the head voxel coordinates. Specifically, for each key voxel, we determine the query position that assigns it the highest attention weight. This operation effectively extracts the most relevant query for each spatial location in the 3D voxels. Finally, we select top-k  voxel coordinates as head voxel coordinates.

\textbf{Tail Voxel Selection.} To address this long-tail distribution problem, \sysname adopts a tail voxel selection module, as shown in the yellow box in \Cref{fig:system_overview}: First, it generates coarse voxel predictions, and then extracts tail voxels equal to the number of head voxels. When the available tail voxels are less than the threshold $K$, the method applies repetition to enhance its representation. Specifically, given the voxel feature $F^{voxel}$, the system first generates a coarse prediction $P^{coarse}$ through a single-layer multi-layer perceptron (MLP) with a softmax activation function, formulating the task as a point-by-point segmentation problem:

\begin{equation}
    \label{equation2}
    \mathbb{P}^{Tail} = \text{MLP}(F^{Tail}),
\end{equation}

we define the coordinate of a 3D voxel as a tail voxel coordinate here. Then, for each voxel $P^{Tail}_{i,j,k}$ at the coordinate $(i,j,k)$, we sort them in descending order according to their tail class probability. Since it is a rough prediction, the tail probability of each voxel may not be accurate, we implement a robust top-K selection strategy. In addition to extracting the voxel with the highest probability of the tail class, we also extract the voxel with the second highest probability of the tail class to ensure that the tail class voxels are robustly extracted without omission. The entire selection process follows the following steps: (1) Extract the voxel $top-1$ with the highest tail probability. (2) If the number is still less than $K$, include the voxel $top-2$ with the second highest probability. (3) Repeat the extraction of these voxels as needed until $K$ samples are reached.

\textbf{Tail Voxel Loss.}
In order to further solve the long tail problem, it is not enough to simply extract the tail voxel. It is necessary to use the corresponding tail ground truth in the target to strongly supervise the extracted tail voxel, forcing the model to pay more attention to the tail voxel. \sysname uses a method similar to HASSC~\cite{wang2024not} to calculate its tail voxel loss. First, \sysname extracting $N$ tail voxels based on their coarse prediction probabilities $P^{Tail} \in \mathbb{R}^{C_{1}\times H\times \times W \times D}$ and corresponding coordinates $V^{Tail}\in \mathbb{R}^{N\times 3}$, where $C_{1}$ represents the number of class and $ H\times \times W \times D$ indicates the spatial shape of voxels. The coordinate $V^{Tail}$ is then used to resample the corresponding features from the fine-grained voxel feature $F^{Tail}\in \mathbb{R}^{C_{2}\times H\times \times W \times D}$ and ground truth $G^{Tail}\in \mathbb{R}^{ H \times W \times D}$, where $C_{2}$ is the number of channels.  The refinement module then processes the concatenated features and predictions through a lightweight MLP to generate refined predictions $P^{Tail}_{Refine} \in \mathbb{R}^{N \times C_{1}}$. The tail voxel loss is calculated as follows:


\begin{equation}
    \label{equation3}
    \mathcal{L}_{TVL} = \text{CE}(P^{Tail}_{Refine}, G^{Tail}),
\end{equation}

where CE denotes the cross-entropy loss and $\mathcal{L}_{TVL}$ represents the tail voxel loss. Therefore, the total loss of \sysname is :
\begin{equation}
    \label{equation4}
    \mathcal{L}_{total} = \mathcal{L}_{baseline} + \mathcal{L}_{TVL},
\end{equation}
where$ \mathcal{L}_{baseline}$ denotes the origin loss of the base line. Adding tail voxel loss can make the model pay more attention to the tail class, thereby improving the overall prediction accuracy

\subsection{Decouple Decoder}
\label{decouple learning}

The most effective way to solve the long-tail problem is to combine decouple learning with data resampling. While this approach works well in 2D tasks, 3D occupancy prediction faces unique challenges: each voxel represents an independent classification task, and resampling the tail class data destroys the geometric structure. Our proposed Sparse Head-Tail-Voxel Construction maintains an equal number of voxels for the head and tail classes while preserving spatial relationships, effectively establishing a voxel resampling framework specifically for 3D spatial reasoning tasks. In addition, our proposed Decouple Decoder enhances decouple learning through label smoothing during classifier optimization, thereby reducing overconfidence in the head class while improving the prediction accuracy of the tail class.

Specifically, our framework implements an iterative refinement process for the selected Head-Tail Voxels. \sysname is designed as a plug-and-play module with broad compatibility across different architectures. The refinement pipeline operates as follows: The identified Head-Tail Voxels are first processed through standard 3D decoder components (cross-attention, self-attention, and feed-forward networks, illustrated in \Cref{fig:system_overview}) to generate refined voxel features. Then, these refined voxel features are then integrated with the encoded voxels through voxel completion. The complete voxel features are finally processed by a segmentation head to produce the semantic voxel predictions. This modular design maintains architectural flexibility while ensuring comprehensive feature refinement through established attention mechanisms and neural network components. Furthermore, we enhance the decoupled learning process through strategic label smoothing during the classifier optimization phase. This modification counteracts model overconfidence in head-class predictions while improving tail-class prediction accuracy, which is shown as following:

\begin{equation}
\label{equation5}
\mathcal{L}_{\text{LS}} = - \sum_{i = 1}^{N} \sum_{c = 1}^{C} \tilde{y}_{i,c} \log(p_{i,c}), \quad \tilde{y}_{i,c} = 
\begin{cases}
1 - \epsilon_c & \text{if } c = y_i \\
\epsilon_c / (C - 1) & \text{otherwise}
\end{cases}
\end{equation}
where $\epsilon_c $ is the smooth factor, $C$ is the number of classes.Our experiments show that adding a small smoothing factor can better balance the category representation while improving the performance of the tail classes.

%% file: chapter/experiment.tex
\section{Experiments}
\subsection{Experiment Setup}
\label{Experiment Setup}

\textbf{Datasets.} We evaluate \sysname on multiple tasks across three benchmark datasets: SemanticKITTI~\cite{behley2019semantickitti} for semantic scene completion, nuScenes-Occupancy~\cite{wang2023openoccupancy} and Occ3D-nuScenes~\cite{tian2023occ3d} for 3D occupancy prediction, and nuScenes~\cite{caesar2020nuscenes} for LiDAR segmentation. \textbf{SemanticKITTI} provides 22 sequences with monocular images, LiDAR point clouds, segmentation labels, and semantic scene completion annotations, using sequences 00-10 (excluding 08) for training, sequence 08 for validation, and sequences 11-21 for testing, with each voxel labeled as either empty or one of 19 semantic classes. \textbf{NuScenes-Occupancy} extends nuScenes with dense 3D semantic occupancy annotations (1 empty class and 16 semantic classes) through an Augmenting and Purifying pipeline, covering 700 training and 150 validation scenes while utilizing the original nuScenes' multi-view images and LiDAR points. \textbf{Occ3D-nuScenes} is a large-scale autonomous driving dataset containing 700 training and 150 validation scenes with 17 semantic categories (16 object classes plus "empty"). The \textbf{nuScenes} dataset contains 1000 driving sequences (20s each) from Boston and Singapore with 2Hz key-frame 3D bounding box annotations. Following Occformer~\cite{zhang2023occformer}, we train \sysname using sparse LiDAR point supervision under the official 700/150/150 train/val/test split for 3D semantic occupancy prediction.

\textbf{Evaluation Metric.}
We report the mean intersection over union (mIoU) for both the semantic scene completion (SSC) and the LiDAR segmentation tasks. For computational efficiency, we report inference time, training memory consumption in SemanticKITTI dataset following ProtoOcc~\cite{oh20253d}, we additionally report training memory, 3D/Overall Latency in nuScenes-Occupancy dataset following the settings from SparseOcc~\cite{tang2024sparseocc}.

\textbf{Implementation Details.}
We integrate \sysname into several state-of-the-art semantic scene completion methods on the SemanticKITTI dataset, including SparseOcc~\cite{tang2024sparseocc} and Symphonize~\cite{jiang2024symphonize}. Following ProtoOcc~\cite{oh20253d}, we evaluate Symphonize-SHTOcc in three configurations: original, base, and small, with query sizes of $128 \times 128 \times 8$, $128 \times 128 \times 16$, and $64 \times 64 \times 8$, respectively. Then, we extend our evaluation to 3D occupancy prediction by integrating \sysname into SparseOcc~\cite{tang2024sparseocc} on the nuScenes-Occupancy dataset~\cite{wang2023openoccupancy} and into COTR~\cite{ma2024cotr} on the Occ3D-nuScenes dataset~\cite{tian2023occ3d}. Furthermore we integrate \sysname to LiDAR segmentation by incorporating it into Occformer~\cite{zhang2023occformer} on the nuScenes dataset. 


%


\subsection{Experimental Results}
\textbf{Semantic Scene Completion.} As shown in \Cref{tab:app_sem_kitti_test}, we report the quantitative comparison of existing state-of-the-art methods for semantic scene completion tasks on SemanticKITTI datsets, the results are evaluated on the test set. In comparison with the lightweight method SparseOcc, \sysname can be further lightweight and improve the performance. What's more, under the query settings of different resolutions proposed by ProtoOCC, the mIOU of \sysname is significantly better than symphonize and ProtoOcc, with an improvement of about $0.3$ compared to ProtoOcc and about $0.7$ compared to symphonize. For efficiency evaluation, We can observe that \sysname can improve the inference speed and save training memory by comparing to all baselines. Specifically, \sysname reduces the inference time by up to $58.6\%$ when compared with ProtoOcc-s, and saves up to $42.2\%$ of training memory when compared with symphonize-o. These results show the superiority of \sysname in saving computational overhead.
\input{table/experiment/SemanticKITTI}

\input{table/experiment/nuscenes-occupancy}

\input{table/experiment/cotr}

\textbf{3D Occupancy Prediction.}
We integrate \sysname into SparseOcc~\cite{tang2024sparseocc} on the nuScenes-Occupancy dataset~\cite{wang2023openoccupancy} in \Cref{table:nusc} and integrate \sysname into COTR~\cite{ma2024cotr} on the Occ3D-nuScenes dataset~\cite{tian2023occ3d} in \Cref{tab:cotr} to evaluate 3D occupancy prediction. The mIOU is improved by 0.7 points and 0.2 points on sparseOcc and COTR respectively. In particular, the training memory and inference time of \sysname are improved by $7\%$ and $25\%$ respectively compared to SparseOcc. It should be emphasized that SparseOcc itself is a lightweight method, \sysname is even lighter and mIOU can be improved.
\input{table/experiment/nuscenes}

\textbf{LiDAR Segmentation.} We assign the voxel predictions on sparse LiDAR points for the semantic segmentation evaluation. As shon in \Cref{tab:lidar_seg}, we report the quantitative comparison of  existing state-of-the-art methods for LiDAR segmentation tasks on nuScenes validation set followed ~\cite{wang2024panoocc}. In this setting, \sysname is integrated into Occformer and implement the R101-DCN as image backbone. The result shows that with SHTOcc, the mIOU of occformer improves $0.6\%$. Additionally, the tail class such as construction vehicle and motorcycle improves significantly, shows the improvement of \sysname in task of LiDAR Segmentation.

\begin{figure}[t]
\centering
    \begin{subfigure}[b]{0.49\linewidth} 
         \centering
         \includegraphics[width=\linewidth]{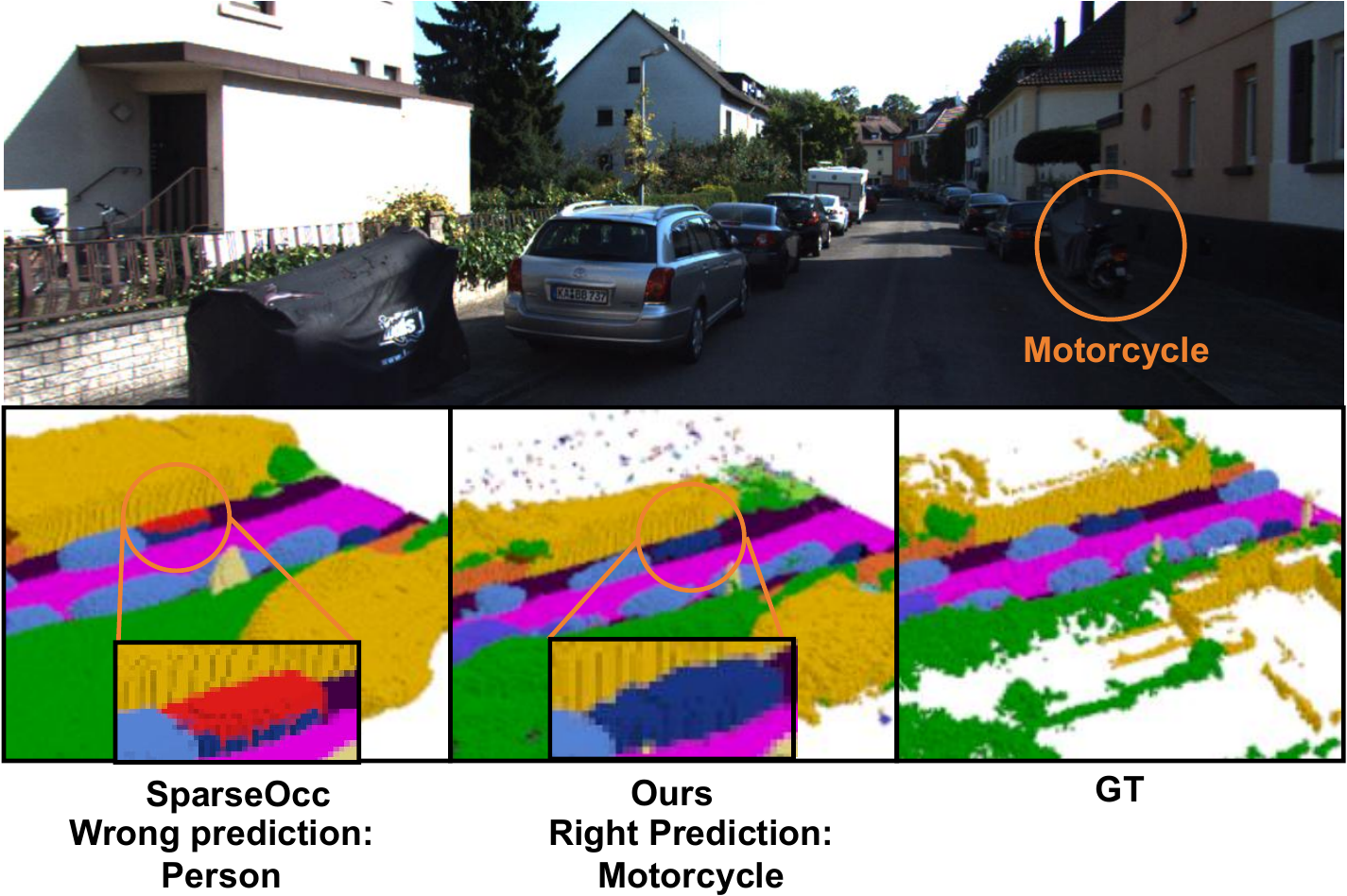} 
         \vspace{-0.2in}
         \caption{Visualization (a). SparseOcc~\cite{tang2024sparseocc} predicts motorcycle as person, while \sysname predicts it correctly. }
         \label{visual1}
    \end{subfigure}
    \hfill 
    \begin{subfigure}[b]{0.49\linewidth}
         \centering
         \includegraphics[width=\linewidth]{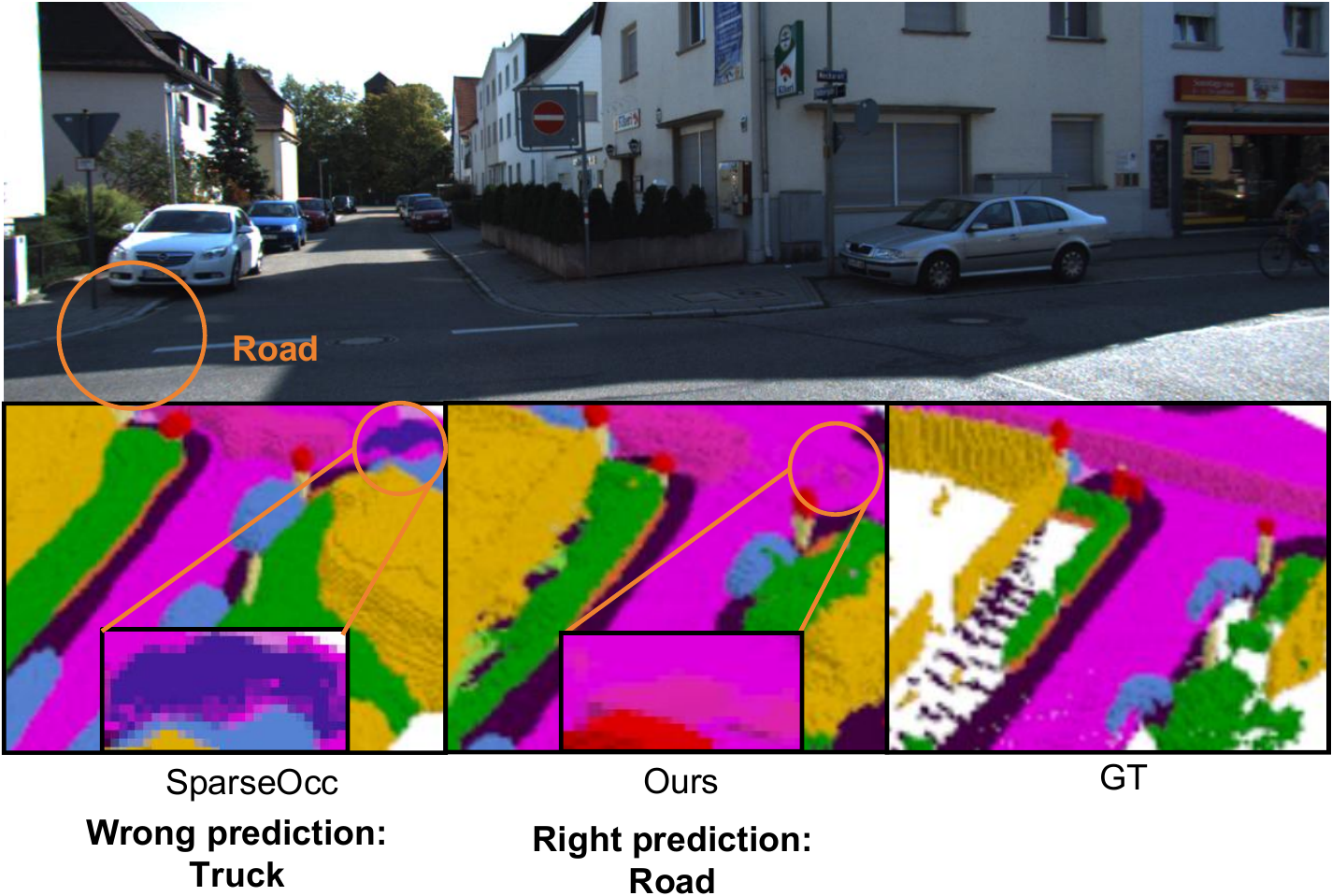}
         \vspace{-0.2in}
         \caption{Visualization (b). SparseOcc~\cite{tang2024sparseocc} predicts road as truck, while \sysname predicts it correctly.}
         \label{visual2}
    \end{subfigure}
        \caption{Visualization of SparseOcc ~\cite{tang2024sparseocc} and SHTOcc. The figure demonstrates that \sysname can achieve more accurate predictions with lower computation cost.}
    \label{visualiazation}
    
\end{figure}

\textbf{Visualization.} \Cref{visualiazation} demonstrates the visualization results of SparseOcc~\cite{tang2024sparseocc} and SHTOcc. SparseOcc selects all non-empty voxels when constructing sparse voxels, but this selection strategy will be affected by many uncritical voxels, resulting in incorrect prediction results. In contrast, \sysname uses the most important voxels that the model focuses on, and the number of selected voxels is much smaller than SparseOcc, but it can make more accurate predictions. As shown in \Cref{visual1} and \Cref{visual2}, motorcycles and roads are predicted as people and trucks, while \sysname predicts correctly.

%% file: table/experiment/SemanticKITTI.tex
\definecolor{car}{rgb}{0.39215686, 0.58823529, 0.96078431}
\definecolor{bicycle}{rgb}{0.39215686, 0.90196078, 0.96078431}
\definecolor{motorcycle}{rgb}{0.11764706, 0.23529412, 0.58823529}
\definecolor{truck}{rgb}{0.31372549, 0.11764706, 0.70588235}
\definecolor{othervehicle}{rgb}{0.39215686, 0.31372549, 0.98039216}
\definecolor{person}{rgb}{1.        , 0.11764706, 0.11764706}
\definecolor{bicyclist}{rgb}{1.        , 0.15686275, 0.78431373}
\definecolor{motorcyclist}{rgb}{0.58823529, 0.11764706, 0.35294118}
\definecolor{road}{rgb}{1.        , 0.        , 1.        }
\definecolor{parking}{rgb}{1.        , 0.58823529, 1.        }
\definecolor{sidewalk}{rgb}{0.29411765, 0.        , 0.29411765}
\definecolor{otherground}{rgb}{0.68627451, 0.        , 0.29411765}
\definecolor{building}{rgb}{1.        , 0.78431373, 0.        }
\definecolor{fence}{rgb}{1.        , 0.47058824, 0.19607843}
\definecolor{vegetation}{rgb}{0.        , 0.68627451, 0.        }
\definecolor{trunk}{rgb}{0.52941176, 0.23529412, 0.        }
\definecolor{terrain}{rgb}{0.58823529, 0.94117647, 0.31372549}
\definecolor{pole}{rgb}{1.        , 0.94117647, 0.58823529}
\definecolor{trafficsign}{rgb}{1.        , 0.        , 0.        }
\definecolor{otherstructure}{rgb}{0.98039215, 0.58823529, 0.}
\definecolor{otherobject}{rgb}{0.19607843, 1.        , 1.        }

\makeatletter
\newcommand{\car@semkitfreq}{3.92}
\newcommand{\bicycle@semkitfreq}{0.03}
\newcommand{\motorcycle@semkitfreq}{0.03}
\newcommand{\truck@semkitfreq}{0.16}
\newcommand{\othervehicle@semkitfreq}{0.20}
\newcommand{\person@semkitfreq}{0.07}
\newcommand{\bicyclist@semkitfreq}{0.07}
\newcommand{\motorcyclist@semkitfreq}{0.05}
\newcommand{\road@semkitfreq}{15.30}
\newcommand{\parking@semkitfreq}{1.12}
\newcommand{\sidewalk@semkitfreq}{11.13}
\newcommand{\otherground@semkitfreq}{0.56}
\newcommand{\building@semkitfreq}{14.1}
\newcommand{\fence@semkitfreq}{3.90}
\newcommand{\vegetation@semkitfreq}{39.3}
\newcommand{\trunk@semkitfreq}{0.51}
\newcommand{\terrain@semkitfreq}{9.17}
\newcommand{\pole@semkitfreq}{0.29}
\newcommand{\trafficsign@semkitfreq}{0.08}
\newcommand{\semkitfreq}[1]{{\csname #1@semkitfreq\endcsname}}

\begin{table*}[ht]
    \caption{\textbf{Quantitative results on SemanticKITTI \texttt{test}.} * denotes the results provided by its official paper.}
    
    \centering
    \newcommand{\clsname}[2]{
        \rotatebox{90}{
            \hspace{-6pt}
            \textcolor{#2}{$\blacksquare$}
            \hspace{-6pt}
            \renewcommand\arraystretch{0.6}
            \begin{tabular}{l}
                #1                                      \\
                \hspace{-4pt} ~\tiny(\semkitfreq{#2}\%) \\
            \end{tabular}
        }}
    \renewcommand{\tabcolsep}{2pt}
    \renewcommand\arraystretch{1.1}
    \scalebox{0.58}
    {
        \begin{tabular}{l|ccc|rrrrrrrrrrrrrrrrrrrr}
            \toprule
            Method                               &
            mIoU                                 &
            Inf. Time           &
            Memory                                 &
            \clsname{road}{road}                 &
            \clsname{sidewalk}{sidewalk}         &
            \clsname{parking}{parking}           &
            \clsname{other-grnd.}{otherground}   &
            \clsname{building}{building}         &
            \clsname{carr}{car}                   &
            \clsname{truck}{truck}               &
            \clsname{bicycle}{bicycle}           &
            \clsname{motorcycle}{motorcycle}     &
            \clsname{other-veh.}{othervehicle}   &
            \clsname{vegetation}{vegetation}     &
            \clsname{trunk}{trunk}               &
            \clsname{terrain}{terrain}           &
            \clsname{person}{person}             &
            \clsname{bicyclist}{bicyclist}       &
            \clsname{motorcyclist}{motorcyclist} &
            \clsname{fence}{fence}               &
            \clsname{pole}{pole}                 &
            \clsname{traf.-sign}{trafficsign}    &
            
            \\
            \midrule
            LMSCNet*~\cite{roldao2020lmscnet}     & 7.07& -&-  & 46.70 & 19.50 & 13.50 & 3.10  & 10.30 & 14.30 & 0.30 & 0.00 & 0.00 & 0.00 & 10.80 & 0.00  & 10.40 & 0.00 & 0.00 & 0.00 & 5.40  & 0.00 & 0.00  \\
            AICNet*~\cite{li2020anisotropic}       & 7.09& -&-  & 39.30 & 18.30 & 19.80 & 1.60  & 9.60  & 15.30 & 0.70 & 0.00 & 0.00 & 0.00 & 9.60  & 1.90  & 13.50 & 0.00 & 0.00 & 0.00 & 5.00  & 0.10 & 0.00 \\
            JS3C-Net*~\cite{yan2021sparse}     & 8.97& -&-  & 47.30 & 21.70 & 19.90 & 2.80  & 12.70 & 20.10 & 0.80 & 0.00 & 0.00 & 4.10 & 14.20 & 3.10  & 12.40 & 0.00 & 0.20 & 0.20 & 8.70  & 1.90 & 0.30 \\
            MonoScene*~\cite{cao2022monoscene}    & 11.08& -&- & 54.70 & 27.10 & 24.80 & 5.70  & 14.40 & 18.80 & 3.30 & 0.50 & 0.70 & 4.40 & 14.90 & 2.40  & 19.50 & 1.00 & 1.40 & 0.40 & 11.10 & 3.30 & 2.10 \\
            TPVFormer*~\cite{huang2023tri}   & 11.26& -&- & 55.10 & 27.20 & 27.40 & 6.50  & 14.80 & 19.20 & 3.70 & 1.00 & 0.50 & 2.30 & 13.90 & 2.60  & 20.40 & 1.10 & 2.40 & 0.30 & 11.00 & 2.90 & 1.50 \\
            OccFormer*~\cite{zhang2023occformer}  & 12.32& -&- & 55.90 & 30.30 & 31.50 & 6.50  & 15.70 & 21.60 & 1.20 & 1.50 & 1.70 & 3.20 & 16.80 & 3.90  & 21.30 & 2.20 & 1.10 & 0.20 & 11.90 & 3.80 & 3.70 \\
            HASSC*~\cite{wang2024not}   & 13.34& -&- & 54.60 & 27.70 & 23.80 & 6.20 & 21.10 & 22.80 & 4.70 & 1.60 & 1.00 & 3.90 & 23.80 & 8.50 & 23.30 & 1.60 & \textbf{4.00} & 0.30 & 13.01 & 5.80 & 5.50 \\
  
            VoxFormer*~\cite{li2023voxformer} & 12.20& -&- & 53.90 & 25.30 & 21.10 & 5.60 & 19.80 & 20.80 & 3.50 & 1.00 & 0.70 & 3.70 & 22.40 & 7.50 & 21.30 & 1.40 & 2.60 & 0.20 & 11.10 & 5.10 & 4.90 \\
           \hline
            SparseOcc*~\cite{tang2024sparseocc}  &11.83 & 51.68ms &13361M& \textbf{55.80} &28.40  & \textbf{31.70}&5.10  &16.00  &21.70 &0.00  & \textbf{1.60} & 0.70 & 3.40 &\textbf{16.40}  &4.10  & \textbf{21.30} & 0.00 &0.00  &0.00  & \textbf{11.70} &\textbf{3.90}  &3.00    \\
            \cellcolor{gray!20}Ours (+SparseOcc)    &\cellcolor{gray!20}\textbf{12.11} & \cellcolor{gray!20}\textbf{50.19ms} &\cellcolor{gray!20}\textbf{11819M}  & \cellcolor{gray!20}53.90 &\cellcolor{gray!20}\textbf{28.60}  &\cellcolor{gray!20}27.70  &  \cellcolor{gray!20}\textbf{7.10} & \cellcolor{gray!20}\textbf{16.10} & \cellcolor{gray!20}\textbf{21.80} &\cellcolor{gray!20}\textbf{2.00}  &\cellcolor{gray!20}1.00 &\cellcolor{gray!20}\textbf{1.60} &\cellcolor{gray!20}\textbf{8.80} &\cellcolor{gray!20}16.20  &\cellcolor{gray!20}\textbf{4.10}   &\cellcolor{gray!20}21.00  &\cellcolor{gray!20}\textbf{1.20} &\cellcolor{gray!20}\textbf{0.40} &\cellcolor{gray!20}\textbf{0.00}  & \cellcolor{gray!20}\textbf{11.70}& \cellcolor{gray!20}3.70&\cellcolor{gray!20}\textbf{3.20} \\
            \hline
            Symphonies-\texttt{o}*~\cite{jiang2024symphonize} & 15.04 &39.79ms&18000M& \textbf{58.40} & \textbf{29.30} & \textbf{26.90} & 11.70 & \textbf{24.70} & 23.60 & 3.20 & \textbf{3.60} & \textbf{2.60} & \textbf{5.60} & 24.20 & 10.00 & \textbf{23.10} & 3.20 & 1.90 & \textbf{2.00} & 16.10 & 7.70 & 8.00 \\
           \cellcolor{gray!20}Ours (+Symphonies-\texttt{o})     &\cellcolor{gray!20}\textbf{15.13}&\cellcolor{gray!20}\textbf{25.38ms}&\cellcolor{gray!20}\textbf{10400M} &\cellcolor{gray!20}57.80  & \cellcolor{gray!20}28.90&\cellcolor{gray!20}26.30&\cellcolor{gray!20}\textbf{12.00}  &\cellcolor{gray!20}23.40  &\cellcolor{gray!20}\textbf{24.60}  &\cellcolor{gray!20}\textbf{3.60}  &\cellcolor{gray!20}3.50  &\cellcolor{gray!20}1.80  & \cellcolor{gray!20}4.60 &\cellcolor{gray!20}\textbf{24.90}  &\cellcolor{gray!20}\textbf{11.00}  &\cellcolor{gray!20}22.50 & \cellcolor{gray!20}\textbf{3.70} &\cellcolor{gray!20}\textbf{3.80}&\cellcolor{gray!20}0.10 &\cellcolor{gray!20}\textbf{17.10} &\cellcolor{gray!20}\textbf{8.70} &\cellcolor{gray!20}\textbf{9.40} \\
            \hline
            ProtoOcc-\texttt{S}*~\cite{oh20253d}   & 14.36&40.80ms&- & 57.20 & \textbf{30.40} & 30.00 & 10.10  & \textbf{24.50} & 22.30 &\textbf{2.80} &2.00 &1.70 & 4.30 & \textbf{24.50} & 8.10  & 23.20 &2.30 & 2.20 & 0.40 &14.60 & 6.20 & 6.30 \\
            Symphonies-\texttt{S}*~\cite{jiang2024symphonize}  & 14.07&23.50ms&14000M & \textbf{57.60} & 29.40 & 29.00 & 10.50 & 24.20 & 22.00 & 2.70 & 2.00 & 1.80 & 4.70 & 23.80 & 7.50 & 22.80 & 2.10 & 1.40 & \textbf{0.70} & 13.80 & 5.60 & 5.70 \\
            
            \cellcolor{gray!20}Ours (+Symphonies-\texttt{S})    &\cellcolor{gray!20}\textbf{14.74}&\cellcolor{gray!20}\textbf{16.91ms}&\cellcolor{gray!20}\textbf{8110M} &\cellcolor{gray!20}57.40  & \cellcolor{gray!20}29.20&\cellcolor{gray!20}\textbf{26.20}&\cellcolor{gray!20}\textbf{11.60}  &\cellcolor{gray!20}23.20  &\cellcolor{gray!20}\textbf{22.70}  &\cellcolor{gray!20}2.10  &\cellcolor{gray!20}\textbf{4.10}  &\cellcolor{gray!20}\textbf{2.20}  & \cellcolor{gray!20}\textbf{6.60} &\cellcolor{gray!20}23.90  &\cellcolor{gray!20}\textbf{9.00}  &\cellcolor{gray!20}\textbf{23.70} & \cellcolor{gray!20}\textbf{4.20} &\cellcolor{gray!20}\textbf{3.50}&\cellcolor{gray!20}0.50 &\cellcolor{gray!20}\textbf{15.30}  &\cellcolor{gray!20}\textbf{7.30}  &\cellcolor{gray!20}\textbf{7.50} \\
            \hline
            ProtoOcc-\texttt{B}*~\cite{oh20253d}   &14.77& 76.20ms&- & 57.90 & 28.60 & \textbf{27.60} & 10.20  & \textbf{24.30} & 24.10 & 3.10 &3.30 & \textbf{3.20} & \textbf{4.70} & 25.00 & 9.20  &22.60 &3.40 & 2.00 & \textbf{1.50} & 15.60 &7.30 & 6.90  \\
            Symphonies-\texttt{B}*~\cite{jiang2024symphonize}  & 14.50& 49.10ms&24100M & 57.00 & 28.00 & 27.50 & 9.10 & 24.10 & 23.70 & 3.70 & 3.60 & 2.40 & 4.40 & 24.60 & 9.90 & 22.50 & 2.90 & 1.70 & 0.50 & 15.70 & 7.50 & 7.00 \\
           
           \cellcolor{gray!20}Ours (+Symphonies-\texttt{B})    &\cellcolor{gray!20}\textbf{15.07}&\cellcolor{gray!20}\textbf{38.75ms} &\cellcolor{gray!20}\textbf{15000M} &\cellcolor{gray!20}\textbf{58.30}  &\cellcolor{gray!20}\textbf{29.00} &\cellcolor{gray!20}26.30&\cellcolor{gray!20}\textbf{10.80}  &\cellcolor{gray!20}22.50  &\cellcolor{gray!20}\textbf{24.40}  &\cellcolor{gray!20}\textbf{4.30}  &\cellcolor{gray!20}\textbf{4.30}  &\cellcolor{gray!20}2.30  &\cellcolor{gray!20}4.00  &\cellcolor{gray!20}\textbf{25.30}  &\cellcolor{gray!20}\textbf{10.60}  &\cellcolor{gray!20}\textbf{22.90} &\cellcolor{gray!20}\textbf{3.80}  &\cellcolor{gray!20}\textbf{2.40}&\cellcolor{gray!20}0.30 &\cellcolor{gray!20}\textbf{17.10} &\cellcolor{gray!20}\textbf{8.50}  &\cellcolor{gray!20}\textbf{9.10} \\
            \bottomrule
        \end{tabular}
    }
    \label{tab:app_sem_kitti_test}
\end{table*}

%% file: table/experiment/nuscenes-occupancy.tex
\definecolor{barrier}{RGB}{112,128,144}
\definecolor{bicycle}{RGB}{220,20,60}
\definecolor{bus}{RGB}{255, 127, 80}
\definecolor{car}{RGB}{255, 158, 0}
\definecolor{const. veh.}{RGB}{233, 150, 70}
\definecolor{motorcycle}{RGB}{255,61,99}
\definecolor{pedestrian}{RGB}{0,0,230}
\definecolor{traffic cone}{RGB}{47,79,79}
\definecolor{trailer}{RGB}{255,140,0}
\definecolor{truck}{RGB}{255,99,71}
\definecolor{drive. suf.}{RGB}{0,207,191}
\definecolor{other flat}{RGB}{175,0,75}
\definecolor{sidewalk}{RGB}{75,0,75}
\definecolor{terrain}{RGB}{112,180,60}
\definecolor{manmade}{RGB}{222,184,135}
\definecolor{vegetation}{RGB}{0,175,0}
\begin{table*}
        \scriptsize
	\setlength{\tabcolsep}{0.0035\linewidth}
	\newcommand{\classfreq}[1]{{~\tiny(\semkitfreq{#1}\%)}}  %
	\centering
    \caption{\textbf{Semantic occpancy prediction results on nuScenes-Occupancy~\cite{wang2023openoccupancy} validation set.}  * represents the results provided by ~\cite{tang2024sparseocc}. \dagger denotes the repreoduced results by its official code.}
   
    \scalebox{0.98}{
	\begin{tabular}{l|c| c c c  |c  c c c c c c c c c c c c c c c c}
 
		\hline
		Method
		& \makecell[c]{Input}
            & \makecell[c]{mIoU}
            & \makecell[c]{Memory}
            & \makecell[c]{3D/Overall \\Latency}
		& \rotatebox{90}{\textcolor{barrier}{$\blacksquare$} barrier} 
		& \rotatebox{90}{\textcolor{bicycle}{$\blacksquare$} bicycle}
		& \rotatebox{90}{\textcolor{bus}{$\blacksquare$} bus} 
		& \rotatebox{90}{\textcolor{car}{$\blacksquare$} car} 
		& \rotatebox{90}{\textcolor{const. veh.}{$\blacksquare$} const. veh.} 
		& \rotatebox{90}{\textcolor{motorcycle}{$\blacksquare$} motorcycle} 
		& \rotatebox{90}{\textcolor{pedestrian}{$\blacksquare$} pedestrian} 
		& \rotatebox{90}{\textcolor{traffic cone}{$\blacksquare$} traffic cone} 
		& \rotatebox{90}{\textcolor{trailer}{$\blacksquare$} trailer} 
		& \rotatebox{90}{\textcolor{truck}{$\blacksquare$} truck} 
		& \rotatebox{90}{\textcolor{drive. suf.}{$\blacksquare$} drive. suf.} 
		& \rotatebox{90}{\textcolor{other flat}{$\blacksquare$} other flat} 
		& \rotatebox{90}{\textcolor{sidewalk}{$\blacksquare$} sidewalk} 
		& \rotatebox{90}{\textcolor{terrain}{$\blacksquare$} terrain} 
		& \rotatebox{90}{\textcolor{manmade}{$\blacksquare$} manmade} 
		& \rotatebox{90}{\textcolor{vegetation}{$\blacksquare$} vegetation}
           
            \\
		\hline\hline
        MonoScene*~\cite{cao2022monoscene} & C  & 6.9& - & - & 7.1  & 3.9  &  9.3 &  7.2 & 5.6  & 3.0  &  5.9& 4.4& 4.9 & 4.2 & 14.9 & 6.3  & 7.9 & 7.4  & 10.0 & 7.6 \\
  
        TPVFormer*~\cite{huang2023tri} &C  &  7.8&  20G & 0.57/0.73s & 9.3  & 4.1  &  11.3 &  10.1 & 5.2  & 4.3  & 5.9 & 5.3&  6.8& 6.5 & 13.6 & 9.0  & 8.3 & 8.0  & 9.2 & 8.2  \\
  
        OpenOccupancy*~\cite{wang2023openoccupancy}  & C & 10.3 & 19G & 0.84/1.22s  &  9.9 & 6.8  & 11.2  & 11.5  & 6.3  & 8.4  & 8.6 & 4.3 & 4.2 & 9.9 & 22.0  & 15.8 & 14.1  & 13.5  & 7.3&10.2  \\
        
        C-CONet*~\cite{wang2023openoccupancy} & C   & 12.8 & 21G & 2.18/2.58s&13.2  & 8.1 &  \textbf{15.4} &  17.2 & 6.3  & 11.2  & 10.0  &  8.3 & 4.7 & 12.1 & 31.4 & 18.8 & 18.7  & 16.3 & 4.8  &8.2 \\
        
        \hline
        SparseOcc\dagger~\cite{tang2024sparseocc} & C  & 13.1 & 15G & 0.06/0.12s & 15.7 & 7.3 & 15.3 & 17.6 & 6.1 & 8.1 & 10.6 & \textbf{9.4} & 5.5 & 12.9 & 30.8 & \textbf{21.7} & 19.1 & 16.1 & 4.9 & 8.6 \\
       
        \cellcolor{gray!20}Ours (+SparseOcc) & C & \textbf{13.8} & \textbf{14G} & \textbf{0.04/0.09s} & \textbf{15.9} & \textbf{8.6} &14.9  & \textbf{18.3} & \textbf{7.3} & \textbf{9.4} & \textbf{10.8} & 9.1 &\textbf{ 6.5} & \textbf{13.2} & \textbf{31.0} & 21.3 & \textbf{20.3} &\textbf{18.4} &\textbf{ 6.1} &\textbf{10.4} \\
        \hline
        \end{tabular}}
        
	\label{table:nusc}
\end{table*}

%% file: table/experiment/cotr.tex
\begin{table}[htbp]
    \centering
    \caption{ Semantic occpancy prediction results on Occ3D-nuScenes~\cite{tian2023occ3d} validation set. * represents the results provided by its official paper.}
     \scalebox{0.82}{\begin{tabular}{l|l|l|c|c|c}
        \toprule
        Method &Venue& Image Backbone &  Epoch & Visible Mask  & mIoU (\%) \\
        \midrule
        MonoScene*~\cite{cao2022monoscene}&CVPR'22   & ResNet-101 &  24 & \ding{55} & 6.1 \\
        OccFormer*~\cite{zhang2023occformer}  &ICCV'23  & ResNet-50  & 24 & \ding{55}  & 20.4 \\
        BEVFormer*~\cite{li2022bevformer}  & ECCV'22 & ResNet-101  & 24 & \ding{55}  & 26.9 \\
        CTF-Occ* ~\cite{tian2023occ3d} & arXiv'23  & ResNet-101  & 24 & \ding{55}  & 28.5 \\
        VoxFormer*~\cite{li2023voxformer} &CVPR'23  & ResNet-101  & 24 & \ding{51}  & 40.7 \\
        SurroundOcc*~\cite{wei2023surroundocc} & ICCV'23  & InternImage-B  & 24 & \ding{51}& 40.7 \\
        FBOcc*~\cite{li2023fb}& ICCV'23  & ResNet-50 & 20 & \ding{51} & 42.1 \\
        PanoOcc*~\cite{wang2024panoocc} &CVPR'24& - & 24 & \ding{51} & 38.1 \\
        ProtoOcc-CVPR*~\cite{oh20253d}&CVPR'25  &-  & 24 & \ding{51} & 39.0 \\
        ProtoOcc-AAAI*~\cite{kim2025protoocc} &AAAI'25 & ResNet-50 & 24 & \ding{51} & 39.6 \\
        STCOcc*~\cite{liao2025stcocc}&CVPR'25  &ResNet-50  & 24 & \ding{51} & 44.6 \\
        COTR+BEVDet3D*~\cite{ma2024cotr} &CVPR'24  & ResNet-50 & 24 & \ding{51}& 44.5 \\
        \midrule
        \cellcolor{gray!20} Ours (+COTR)  & \cellcolor{gray!20}- &  \cellcolor{gray!20}ResNet-50 & \cellcolor{gray!20} 24 &  \cellcolor{gray!20}\ding{51}&  \cellcolor{gray!20}44.7 \\
        
        \bottomrule
    \end{tabular}}
    
    \label{tab:cotr}
\end{table}

%% file: table/experiment/nuscenes.tex
\definecolor{nbarrier}{RGB}{255, 120, 50}
\definecolor{nbicycle}{RGB}{255, 192, 203}
\definecolor{nbus}{RGB}{255, 255, 0}
\definecolor{ncar}{RGB}{0, 150, 245}
\definecolor{nconstruct}{RGB}{0, 255, 255}
\definecolor{nmotor}{RGB}{200, 180, 0}
\definecolor{npedestrian}{RGB}{255, 0, 255}
\definecolor{ntraffic}{RGB}{255, 240, 150}
\definecolor{ntrailer}{RGB}{135, 60, 0}
\definecolor{ntruck}{RGB}{255, 0, 0}
\definecolor{ndriveable}{RGB}{213, 213, 213}
\definecolor{nother}{RGB}{139, 137, 137}
\definecolor{nsidewalk}{RGB}{75, 0, 75}
\definecolor{nterrain}{RGB}{150, 240, 80}
\definecolor{nmanmade}{RGB}{160, 32, 240}
\definecolor{nvegetation}{RGB}{0, 175, 0}

\begin{table*}[ht]
	\scriptsize  
 	\setlength{\tabcolsep}{0.0040\linewidth}
	  \caption{LiDAR semantic segmentation results on nuScenes validation set. * denotes the results provided by the official paper.}
	\newcommand{\classfreq}[1]{{~\tiny(\nuscenesfreq{#1}\%)}}  %
    \begin{center}
         \scalebox{0.92}{
	\begin{tabular}{l|c|c|c | c c c c c c c c c c c c c c c c}
		\toprule
		Method
		& \makecell{Input \\ Modality}& \makecell{Image \\ Backbone} & mIoU
		& \rotatebox{90}{\textcolor{nbarrier}{$\blacksquare$} barrier}
		
		& \rotatebox{90}{\textcolor{nbicycle}{$\blacksquare$} bicycle}
		
		& \rotatebox{90}{\textcolor{nbus}{$\blacksquare$} bus}

		& \rotatebox{90}{\textcolor{ncar}{$\blacksquare$} car}

		& \rotatebox{90}{\textcolor{nconstruct}{$\blacksquare$} const. veh.}

		& \rotatebox{90}{\textcolor{nmotor}{$\blacksquare$} motorcycle}

		& \rotatebox{90}{\textcolor{npedestrian}{$\blacksquare$} pedestrian}

		& \rotatebox{90}{\textcolor{ntraffic}{$\blacksquare$} traffic cone}

		& \rotatebox{90}{\textcolor{ntrailer}{$\blacksquare$} trailer}

		& \rotatebox{90}{\textcolor{ntruck}{$\blacksquare$} truck}

		& \rotatebox{90}{\textcolor{ndriveable}{$\blacksquare$} drive. suf.}

		& \rotatebox{90}{\textcolor{nother}{$\blacksquare$} other flat}

		& \rotatebox{90}{\textcolor{nsidewalk}{$\blacksquare$} sidewalk}

		& \rotatebox{90}{\textcolor{nterrain}{$\blacksquare$} terrain}

		& \rotatebox{90}{\textcolor{nmanmade}{$\blacksquare$} manmade}

		& \rotatebox{90}{\textcolor{nvegetation}{$\blacksquare$} vegetation}

		\\
		\midrule

        RangeNet++*~\cite{milioto2019rangenet++} & LiDAR & - & 65.5 & 66.0 & 21.3 & 77.2 & 80.9 & 30.2 & 66.8 & 69.6 &  52.1 & 54.2 & {72.3} & {94.1} & 66.6 & 63.5 & 70.1 & 83.1 & 79.8 \\
		
		PolarNet*~\cite{zhang2020polarnet} & LiDAR &- & 71.0 & 74.7 & 28.2 & 85.3 & 90.9 & 35.1 & 77.5 & 71.3 & 58.8 & 57.4 & 76.1 & 96.5 & 71.1 & 74.7 & 74.0 & 87.3 & 85.7  \\
		
		Salsanext*~\cite{cortinhal2020salsanext} & LiDAR &- & 72.2 & 74.8 & 34.1 & 85.9 & 88.4 & 42.2 & 72.4 & 72.2 & 63.1 & 61.3 & 76.5 & 96.0 & 70.8 & 71.2 & 71.5 & 86.7 & 84.4 \\
		
		Cylinder3D++*~\cite{zhu2021cylindrical} & LiDAR &- & 76.1 & 76.4 & 40.3 & 91.2 & 93.8 & 51.3 & 78.0 & 78.9 & 64.9 & 62.1 & 84.4 & 96.8 & 71.6 & 76.4 & 75.4 & 90.5 & 87.4 \\
        RPVNet~\cite{xu2021rpvnet} & LiDAR  &- & 77.6 & 78.2 & 43.4 & 92.7 & 93.2 & 49.0 & 85.7 & 80.5 & 66.0 & 66.9 & 84.0 & 96.9 & 73.5 & 75.9 & 76.0 & 90.6 & 88.9 \\
			\midrule	
		
        BEVFormer-Base*~\cite{li2022bevformer} & Camera &R101-DCN & 56.2  & 54.0 & 22.8 & 76.7 & 74.0 & 45.8 & 53.1 & 44.5 & 24.7 & 54.7 & 65.5 & 88.5 & 58.1 & 50.5 & 52.8 & 71.0 & 63.0  \\
		
		TPVFormer-Base*~\cite{huang2023tri}  & Camera & R101-DCN &68.9  & 70.0 & 40.9 & 93.7 & 85.6 & 49.8 & 68.4 & 59.7 & 38.2 & 65.3 & 83.0 & 93.3 & 64.4 & 64.3 & 64.5 & 81.6 & 79.3  \\ %
         \midrule
         Occformer*~\cite{zhang2023occformer} &  Camera & R101-DCN &70.4  & \textbf{70.3} & \textbf{43.8} & 93.2 & \textbf{85.2} & 52.0 & 59.1 & \textbf{67.6} & \textbf{45.4} & 64.4 & 84.5 & \textbf{93.8 }& \textbf{68.2} & 67.8 & 68.3 & \textbf{82.1} &\textbf{ 80.4} \\ %
        \cellcolor{gray!20}Ours (+Occformer) &  \cellcolor{gray!20}Camera & \cellcolor{gray!20}R101-DCN &\cellcolor{gray!20}\textbf{71.0}  & \cellcolor{gray!20}\textbf{70.3} &\cellcolor{gray!20} 43.2 & \cellcolor{gray!20}92.0 & \cellcolor{gray!20}84.3 &\cellcolor{gray!20} \cellcolor{gray!20}\textbf{56.0} & \cellcolor{gray!20}73.5 & \cellcolor{gray!20}64.8 &\cellcolor{gray!20} 43.1 & \cellcolor{gray!20}\textbf{65.0} & \cellcolor{gray!20}\textbf{85.0 }& \cellcolor{gray!20}93.4 &\cellcolor{gray!20} 66.5 &\cellcolor{gray!20} \textbf{68.0} & \cellcolor{gray!20}\textbf{68.5} & \cellcolor{gray!20}81.8 & \cellcolor{gray!20}80.1 \\

		\bottomrule
	\end{tabular}}
    \end{center}
   
	\label{tab:lidar_seg}
\end{table*}

%% file: chapter/ablation_study.tex
\section{Ablation Study}
In this section we evaluate the effectiveness of different components and different tail class groups, more ablation study will be shown in supplementary materials.


\textbf{Ablation on different components.} In order to gain a deeper understanding of the functions of each part, we evaluated the performance improvement of each module in \Cref{different_stage}. After using the head voxel selection module instead of dense prediction, along with the reduction in computational overhead, the performance also decreased slightly, and mIOU dropped by about $0.5$ points. After adding the tail voxel selection module, mIOU increased a little bit, indicating that the tail class voxel is indeed ignored by the model. Adding this part helps the model focus on the voxels of the tail class. After adding Tail voxel loss, mIOU increased by about $0.2$ points, indicating that tail voxel loss is also helpful in guiding the model to pay attention to the tail class voxels. After adding decouple learning, mIOU continued to increase by $0.3$ points, which also highlights the contribution of decouple learning to the long-tail distribution problem.



\input{table/ablation_study/different_stage}

%% file: table/ablation_study/different_stage.tex
\begin{table}[ht]
    \centering
    \renewcommand{\tabcolsep}{4pt}
    \renewcommand\arraystretch{1.1}
    \caption{Ablation study on different modules.}
    \scalebox{0.92}{
        \begin{tabular}{ccccc|c>{\columncolor{gray!20}}c}
            \toprule
            Baseline.       & Head Voxel Selec. & Tail Voxel Selec. & Tail Voxel Loss. & Decouple Learning      & Memory                    & {\cellcolor{gray!20}mIoU}                     \\
            \midrule
               $\checkmark$         &            &            &            &            &           18000M       &14.59                                     \\
            $\checkmark$ &     $\checkmark$       &            &            &            &    10400M           &      14.05                                 \\
             $\checkmark$ & $\checkmark$ & $\checkmark$ &            &            &         10400M    &                    14.18              \\
             $\checkmark$ & $\checkmark$ &        $\checkmark$    & $\checkmark$ &           &             10400M     &  14.40                                   \\
             $\checkmark$ &$\checkmark$ &        $\checkmark$    &$\checkmark$ & $\checkmark$ &               4870M  &     14.75                                 \\
            
            \bottomrule
        \end{tabular}}
    
    \label{different_stage}
\end{table}

%% file: chapter/conclusion.tex
\section{Conclusion}
In this paper, we introduce SHTOcc, a plug-and-play method for saving computational overhead without losing performance. \sysname extracts key voxels and updates them based on the attention mechanism and the distribution of the tail class voxels, and introduces decouple learning to eliminate the influence of the long-tail distribution. Extensive experiments show that \sysname can improve prediction accuracy while saving computational overhead. We expect \sysname to inspire future research and contribute to the progress of autonomous driving and 3D perception.